%% file: bmvc_final.tex
\documentclass{bmvc2k}
\usepackage[colorinlistoftodos]{todonotes}

\usepackage{amsmath,amsfonts,amssymb,amsthm}
\usepackage{mathtools}
\usepackage{booktabs}
\title{TAGAN: Tonality-Alignment Generative Adversarial Networks for Realistic Hand Pose Synthesis}

\addauthor{Liangjian Chen}{liangjc2@ics.uci.edu}{1}
\addauthor{Shih-Yao Lin}{shihyaolin@tencent.com}{2}
\addauthor{Yusheng Xie}{yushengxie@tencent.com}{2}
\addauthor{Hui Tang}{tanghui@tencent.com}{2}
\addauthor{Yufan Xue}{robinxue@tencent.com}{2}
\addauthor{Yen-Yu Lin}{yylin@citi.sinica.edu.tw}{3}
\addauthor{Xiaohui Xie}{xhx@ics.uci.edu}{1}
\addauthor{Wei Fan}{davidwfan@tencent.com}{2}
\addinstitution{
University of California, Irvine
}
\addinstitution{
Tencent Medical AI Lab\\
}
\addinstitution{
Academia Sinica
}
\runninghead{L. Chen \MakeLowercase{et al.}}{Tonality-Alignment Generative Adversarial Networks}

\def\eg{\emph{e.g}\bmvaOneDot}

\def\etal{\emph{et al}\bmvaOneDot}

\DeclareMathOperator{\E}{\mathbb{E}}
\newcommand{\eqnref}[1]{Eq.~(\ref{eqn:#1})}

\DeclarePairedDelimiter{\norm}{\lVert}{\rVert}
\newcommand{\ie}{{\em i.e., }\normalfont}
\begin{document}

\maketitle
\input{hpe_Abstract.tex}
\input{hpe_Intro.tex}

\input{hpe_related_work.tex}

\input{hpe_Main.tex}

\input{hpe_setting.tex}
\input{hpe_results.tex}

\input{hpe_Conclusion.tex}

\paragraph{Acknowledgement.}
This work was supported in part by Ministry of Science and Technology (MOST) under grants 107-2628-E-001-005-MY3 and 108-2634-F-007-009.

\bibliography{egbib}
\end{document}

%% file: hpe_Abstract.tex
\begin{abstract}
$3$D hand pose estimation from a single RGB image is important but challenging due to the lack of sufficiently large hand pose datasets with accurate 3D hand keypoint annotations for training. 
In this work, we present an effective method for generating realistic hand poses, and show that existing algorithms for hand pose estimation can be greatly improved by augmenting training data with the generated hand poses, which come naturally with ground-truth annotations. 
Specifically, we adopt an augmented reality simulator to synthesize hand poses with accurate $3$D hand-keypoint annotations. 
These synthesized hand poses look unnatural and are not adequate for training.  
%
%
To produce more realistic hand poses, we propose to blend each synthetic hand pose with a real background and develop {\em tonality-alignment generative adversarial networks} (TAGAN), which align the tonality and color distributions between synthetic hand poses and real backgrounds, and can generate high-quality hand poses.
%
TAGAN is evaluated on the {\em RHP}, {\em STB}, and {\em CMU-PS} hand pose datasets. 
With the aid of the synthesized poses, our approach performs favorably against the state-of-the-art methods in both $2$D and $3$D hand pose estimation.

%
\end{abstract}

%% file: hpe_Intro.tex
\section{Introduction}

Estimating hand poses from monocular RGB images has drawn increasing attention because it is essential to many applications such as virtual and augmented reality~\cite{mueller_cvpr18}, and human computer interaction~\cite{sridhar_chi15}. 
%
%
It has gained significant progress \cite{Garcia_cvpr18,mueller_cvpr18} owing to the fast development of {\em deep neural networks} (DNN).
These DNN-based methods learn hand representations and estimate poses jointly.
Despite effectiveness, DNN-based methods highly rely on a vast amount of training data. 
However, it is expensive to collect all hand poses of interest with manual hand-keypoint annotations for training.

\begin{figure}[t]
	\begin{center}
		\includegraphics[width=3.0in]{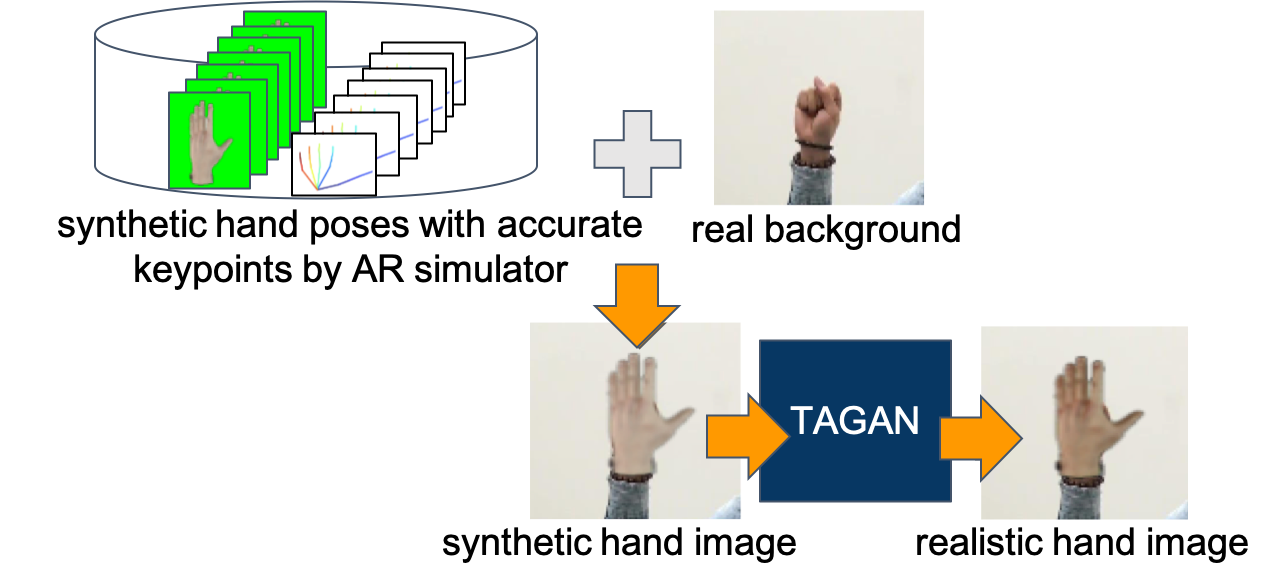}
	\end{center}
	\label{fig:method_overview}
	\vspace{-0.25in}
    \caption{
    Overview of our method for realistic hand image synthesis. We blend a synthetic pose by an AR simulator with real background to yield a synthetic hand image, which is then fed to the proposed TAGAN to produce a more realistic hand image.}
\end{figure}

%
Synthesizing training data has been a feasible way to tackle the lack of training data.
Recent studies, \eg \cite{mueller_cvpr18,zimmermann_iccv17}, have adopted {\em augmented reality} (AR) simulators to generate large-scale training examples. 
In this way, plenty hand images with various poses, skin textures, and lighting conditions can be systematically synthesized.
Moreover, accurate hand-keypoint annotations of these synthesized hand images are also available. 
Training with such synthetic images may not result in a much improved hand pose estimator because of the dissimilarity between the real and synthetic data.
In this work, we suggest blending a synthetic hand pose (foreground) image with a real background image so that the blended images are realistic enough to serve as high-quality training data.

We are aware of the dissimilarity between synthetic hand pose images and real background images in styles and appearances.
Thus, we present a GAN-based method, {\em tonality-alignment generative adversarial networks} (TAGAN), to eliminate the dissimilarity.
TAGAN employs the image-to-image translation technique based on {\em conditional GAN} (CGAN)~\cite{isola_cvpr17}, where extra shape features serve as the input to GAN and constrain the object shape in the synthesized photo. 
%
In addition to the shape constraint, a tonality-alignment loss in TAGAN is designed to align the color distributions tonality of the input and generated images.
It turns out that the hand pose images can be better blended into the background images, resulting in more realistic hand pose images.
The hand pose estimator is then considerably improved by using the generated hand pose images as the augmented training data.

Figure~\ref{fig:method_overview} gives the overview of the proposed method.
The main contribution of this work is three-fold:
First, we propose to fuse synthetic hand poses and real background images so that the resulting synthesized hand images can be more realistic.
Second, we present TAGAN which performs conditional adversarial learning and seamlessly blends synthetic hand poses into real backgrounds.
%
%
%
Third, we demonstrate that existing pose estimators trained with the generated hand pose data gain significant improvements over the current state-of-the-arts on both $2$D and $3$D datasets.

%% file: hpe_related_work.tex
\section{Related Work}


{\flushleft {\bf Data Augmentation via Simulator.}}
%
%
Recent work, \eg \cite{khan_cvpr18}, for hand pose estimation has trained the models on synthetic training data. 
In~\cite{zimmermann_iccv17}, a synthetic hand pose dataset is generated by an open source simulator, and
serves as augmented training data to improve pose estimator learning.
However, the synthetic hand images produced by the AR simulator look artificial, leading to limited performance gains.
To address this issue, recent work, \eg \cite{shrivastava_cvpr17,mueller_cvpr18}, leverages adversarial learning to enhance the quality of synthetic hand images.  

{\flushleft {\bf Data Augmentation via Adversarial Learning.}}
Generating realistic images by using {\em generative adversarial networks} (GAN)~\cite{goodfellow_nips14,shrivastava_cvpr17}  has been a research trend.  
Isola~\etal propose {\em Pix2Pix Net}~\cite{pix2pix_cvpr17} to learn a mapping from a sketch to a realistic image, \eg transferring a car sketch to a car image. 
Unlike GAN requiring paired training data, {\em CycleGAN}~\cite{CycleGAN_iccv17} employs cycle-consistent adversarial networks for translating images from a source domain to a target domain with unpaired examples.
To increase the amount of training data, Shrivastava~\etal present {\em SimGAN}~\cite{shrivastava_cvpr17}, which employs simulated and unsupervised learning to improve the realism of the output of a simulator with unlabeled real data.
However, the simulator's data include only objects, ignoring background scenes. 
Thus, the resulting synthetic images are filled objects, but the background information is usually crucial in practice. 
In this work, we explore techniques that directly regularize the foreground (hand) and the background (natural scenes where the hand appears \cite{zhang_icip16,ionescu_pami14}).
%




{\flushleft {\bf Vision-based Hand Pose Estimation.}}
Hand pose estimation has drawn increasing attention for decades \cite{abdi_bmvc18,baek_cvpr18,huang_bmvc18,Garcia_cvpr18,yuan_cvpr18,wan_cvpr18,ge_cvpr18,moon_cvpr18,pandey_eccv18,ye_eccv18,zhou_eccv18,ge2019handshapepose,WCRSF,boukhayma20193d,tekin2019h+,yang2018disentangling}.
Research efforts can be categorized by their input data forms, which primarily include $2$D RGB images and $3$D RGBD images with depth information. 
Recent progress has tried to estimate the $3$D hand pose from a monocular RGB image. 
For example, Oikonomidis~\etal \cite{oikonomidis_bmvc11,oikonomidis_cvpr12} propose a hand tracking approach based on {\em particle swarm optimization}.
Simon~\etal \cite{simon_cvpr17} adopt multiview bootstrapping to calculate hand keypoints from RGB images. 
Zimmermann and Brox~\cite{zimmermann_iccv17} propose a $3$D pose regression net, enabling $3$D hand pose estimation from an RGB image.

{\flushleft {\bf Domain Adaption via Adversarial Learning.}}
Domain adaption has been introduced by Saenko~\etal~\cite{saenko2010adapting} for pairwise metric transforming and can be developed by the study of visual dataset bias~\cite{torralba2011unbiased}. For domain adaption, models are often designed to capture the invariant patterns between two different data distributions so that they can perform well cross domains. Hoffman~\etal~\cite{hoffman2017cycada} adapt the CycleGAN loss with the task loss to further improve the results of image translation.

%% file: hpe_Main.tex
\section{Methodology}


This section describes our approach to hand pose image generation. 
We first explain how GAN and conditional GAN are applied to this problem, then depict synthetic hand image generation, and finally specify how our approach works to improve the synthetic images.


\subsection{GAN and Conditional GAN}


GAN learns a mapping from a random noise vector $z$ to its generated image $y$, \ie $G: z \rightarrow y$, where $G$ is the generator. 
The {\em conditional GAN} (CGAN)~\cite{mirza_arxiv14} is an extension of GAN.
The inputs to CGAN can be augmented with additional conditions so that CGAN can leverage the additional conditions to constrain the output image $y$. 
The conditions can be specified in the form of extra inputs to both the generator network and the discriminator network. 


CGAN-based methods can be applied to image-to-image translation. 
In the representative work {\em pix2pix net}~\cite{isola_cvpr17}, the condition is used to make the object shape in the output image similar to the additional input shape map $x_{s}$, which is pre-computed by applying the edge detector HED~\cite{xie_cvpr15} to the image $x$.
As the additional input, $x_s$ is fed to both the generator and the discriminator.
In this way, the condition $x_s$ and the latent space representation $z$ are transformed into a joint hidden representation. 
CGAN learns a mapping from $x_s$ (inferred from $x$) and $z$ to the generated output $y$, $G$: $\{x,z\}\rightarrow y$. 
The CGAN objective can be formulated as
\begin{equation}
    \mathcal{L}_{CGAN}(G,D) = \E_{x_s,y}  [\log D(x_s,y)] 
+ \E_{x_s,z} [\log (1-D(x_s,G(x_s,z)))],
 \label{eqn:CGANs}
\end{equation}
where the discriminator $D$ aims to distinguish the data generated by $G$ from real data. Yet, the generator $G$ generates the data to not only fool $D$ but also fulfill the input condition.
In~\eqnref{CGANs}, the generator $G$ minimizes the differences between the real images and the generated images while $G$'s adversary, $D$, tries to learn a discriminating function to maximize such differences.
In addition, the shape of output $y$ is constrained by $x_s$. 
Thus, $G$ is optimized via
\begin{equation}
    G^{*} = \arg\min_{G}\max_{D}\mathcal{L}_{CGAN}(G,D)+\lambda \cdot \mathcal{L}_{S}(G),
\end{equation}
where $\mathcal{L}_{s}$ and $\lambda$ are the shape loss and its weight, respectively. 
The shape loss can be calculated by using $L_1$ distance to reduce the unfavorable effect of blurring, and is defined by
\begin{equation}
     \mathcal{L}_{S}(G)=\E_{x_s,y,z}[ \norm{y-G(x_s,z)}_{1}].
\end{equation}

Although the existing work, Pix2pix Net is able to contain the shape of the generated object by using the input shape map, it does not take color consistency between the object and background into account. Hence, its generated images might look unnatural. 


\begin{figure}[t]
	\begin{center}
		\includegraphics[height=1.1in]{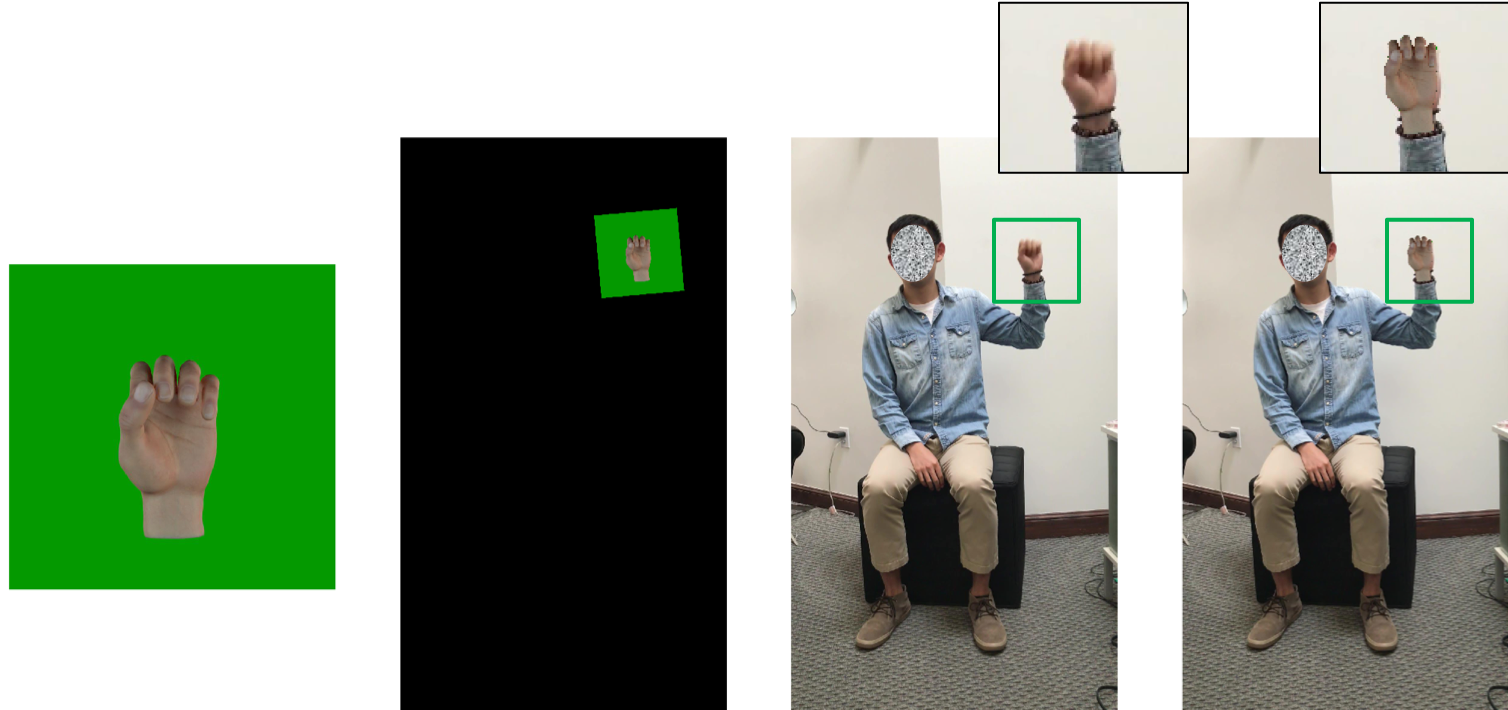}
	\end{center}
	\vspace{-0.2in}
	\caption{
	Hand image synthesis. From left to right: 1) a hand image $\mathbf{u}$ in our synthetic dataset, 2) its representation with an affine transformation $g(\mathbf{u})$, 3) target image (enclosed by the green box) $\mathbf{v}$ via hand detection, and 4) the synthesized hand image obtained by applying TAGAN to blend the synthetic hand pose into the real background image. 
	}
	\label{fig:affine}
\end{figure}

\subsection{Synthetic Hand Image Generation}


Existing hand pose datasets are not large enough to stably learn a deep network for hand pose estimation.
Moreover, manual hand-keypoint annotation is expensive and labor-intensive. 
Also, the annotated hand-keypoints are still error-prone and often not accurate enough.
To address the quantitative and qualitative issues of hand pose training data, we firstly adopt an open-source AR simulator to produce large-scale {\em synthetic hand pose images} with {\em accurate $2$D/$3$D hand-keypoint labels}, which cover common and feasible hand poses by observing a group of subjects for a period of time.
However, these synthetic hands look unnatural and cannot serve as training data.

We also collect a large-scale, daily-life, and unlabeled hand gesture videos performed by some subjects. 
Each image in these videos consists of {\em real hand(s) and background}.
See the third image in Figure~\ref{fig:affine} as an example. 
We detect the hand region in the image using the pose estimation toolkit, OpenPose Library~\cite{shrivastava_cvpr17}.
However, the estimated poses are not good enough to serve as the training data.
Thus, we propose to {\em match} the hand pose images in the AR and real datasets. 
In this way, the accurate keypoints in the AR dataset and real backgrounds in the real dataset can complement each other, and the proposed TAGAN can produce more realistic hand poses with accurate keypoints.

%
%
%
%
%
%
%
%
%
%

Given a hand image with the estimated pose $\mathbf{v}$ in the real dataset, our goal is to synthesize a hand image with its pose consistent with $\mathbf{v}$.
To find the best match, we use $\mathbf{v}$ as a query to the AR dataset generated by the AR simulator, which covers millions of hand poses with accurate keypoint annotations. 
For each candidate hand pose $\mathbf{u}$ in the AR dataset, its similarity to the target pose $\mathbf{v}$ is defined as
\begin{equation}
K(\mathbf{u},\mathbf{v}) = \langle f\circ g(\mathbf{u}), f(\mathbf{v})\rangle /(\Vert f\circ g(\mathbf{u}) \Vert \Vert f(\mathbf{v}) \Vert), \label{eqn:similarity}
\end{equation}
where function $g$ is the affine transformation with which the transformed candidate pose $g(\mathbf{u})$ can best match $\mathbf{v}$, and function $f$ is the feature representation of a pose.
In this work, each hand pose is expressed as the concatenation vector of its $21$ $2$D keypoints, \eg $\mathbf{u} = [u_{x1},u_{y1},u_{x2},u_{y2},\ldots,u_{x21},u_{y21}] \in \mathbb{R}^{42}$. 
The feature representation $f$ of a hand pose is the ordered collection of pair-wise keypoint differences along the $x$ and $y$ axes, \ie
\begin{equation}
f(\mathbf{u}) = [ \ldots, u_{xi}-u_{xj} , u_{yi}-u_{yj} ,\ldots ], \mbox{ for } 1 \leq i < j \leq 21. \label{eqn:pose_representation}
\end{equation}

The candidate pose $\mathbf{u}^{\ast} = \arg\max_{\mathbf{u}}K(\mathbf{u},\mathbf{v})$ is selected from the dataset yielded by the AR simulator.
We superpose pose $\mathbf{u}^{\ast}$ over the scene covering $\mathbf{v}$, and apply the proposed TAGAN to better blend the selected pose into the background scene to produce a more realistic hand image. 
Figure~\ref{fig:affine} summarizes the process. 
The proposed TAGAN is elaborated below.


\begin{figure}[t]
	\begin{center}
		\includegraphics[width=0.6\textwidth]{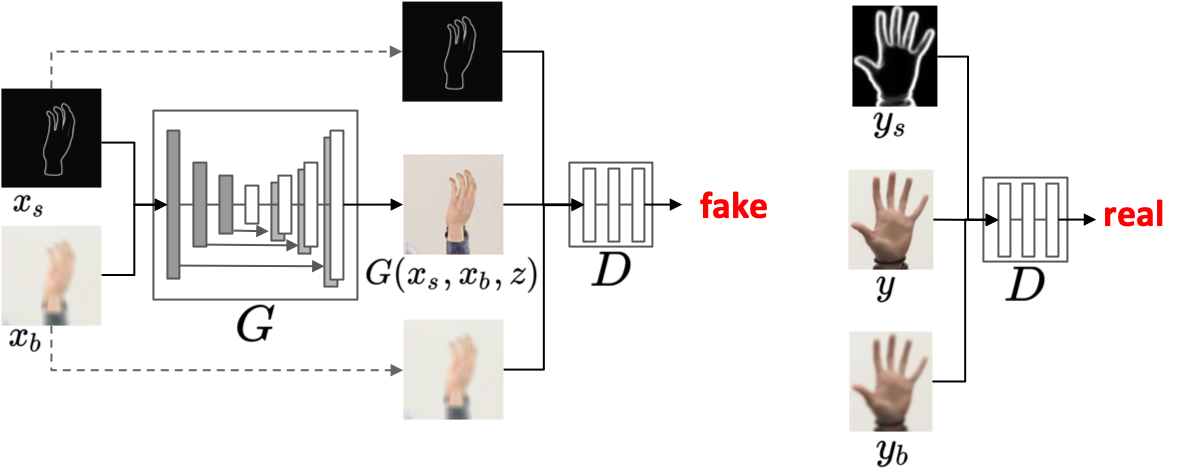}
	\end{center}
	\vspace{-0.25in}
	\caption{
	TAGAN derives a mapping from the shape map $x_s$ and the color map $x_b$ to the generated image $G(x_s,x_b,z)$. 
	The generator $G$ learns to produce realistic images to fool the discriminator $D$ by blending a synthetic hand pose with a real background image, while the discriminator $D$ aims to separate the fake (synthetic) images from the real images. 
	%
	%
	%
	}
	

	\label{fig:core_idea}
\end{figure}

	%
	%
	%

\subsection{Tonality-Alignment GAN}


Although data augmentation using AR simulators can relieve the lack of training data, the background of the synthesized images is artificial.
The background tonality and color distributions between the synthetic and real hand poses are inconsistent. 
These issues make the synthetic hand poses less qualified as training data.
Inspired by the pix2pix net~\cite{pix2pix_cvpr17} that leverages the shape map to constrain the output image, we propose {\em tonality-alignment} GAN (TAGAN) to take the color distribution and shape features into account.


Given a superposed image $x$, we utilize its blurred counterpart $x_b$ and shape map $x_s$ as the color and shape reference, respectively. 
The blurred counterpart in our system is derived by applying an average filter to $x$, while the shape map $x_s$ is obtained by using the HED detector. 
For the real image $y$, we adopt the same scheme to extract the shape map $y_s$ and color maps $y_b$.
In TAGAN, the shape map $x_s$ and color map $x_b$ are fed to both the generator and the discriminator as additional input layers such that the $x_s$, $x_b$ and the output $G(x_s, x_b, z)$ are transformed into a joint hidden representation. 
Figure~\ref{fig:core_idea} illustrates the proposed TAGAN.


During training, the TAGAN learns a mapping from $x_s$, $x_b$ and a random vector $z$ to the generated output $y$, \ie $G$:$\{x_s,x_b, z\}\rightarrow y$.
The objective of TAGAN is designed as
\begin{align}
   \mathcal{L}_{TAGAN}(G,D) = &\E_{y_s,y_b,y}  [\log D(y_s,y_b,y)]+\E_{x_s,x_b,z} [\log (1-D(x_s,x_b,G(x_s,x_b,z)].
\end{align}
The generator $G$ in TAGAN is optimized via
\begin{equation}
    G^{*} = \arg\min_{G}\max_{D}\mathcal{L}_{TAGAN}(G,D)+\mathcal{L}_{TA}(G,x_s,x_b),
\end{equation}
where $\mathcal{L}_{TA}$ is the loss function for enforcing the shape similarity between $x_s$ and $y$ as well as the color consistency between $x_b$ and $y$. The loss is defined by
\begin{equation}
 \mathcal{L}_{TA}(G)=\E_{x_s,x_b,y,z}[\lambda_{1} \cdot D_{c}(x_b,x_s,z,y)+\lambda_{2} \cdot D_{s}(x_b,x_s,z,y)],
\end{equation}\
where $D_c( \cdot )$ and $D_s( \cdot )$ denote the color and shape distance functions, respectively. Constants $\lambda_{1}$ and $\lambda_{2}$ are the weights. The shape distance function $D_s$ is expressed as 
\begin{equation}
    D_{s}(x_b,x_s,z,y) = \norm{y-G(x_s,x_b,z)}_{1}.
\end{equation}
In addition to the shape condition, we design a tonality-alignment loss to align the color distributions of the input and the generated images via defining $D_c$ as
\begin{equation}
    D_{c}(x_b,x_s,z,y) = -\sum_{i}h_{g}(i)\log \Big(\dfrac{h_{y}(i)}{h_{g}(i)}\Big),
    \label{eqn:DC}
\end{equation}
where $h_{y}$ and $h_{g}$ are the color histograms of $y$ and $G(x_b,x_s,z)$, respectively.
Thereby, $D_{c}(\cdot)$ in \eqnref{DC} is the Kullback-Leibler divergence between the two histograms. 
%
To train our TAGAN, we collect a real unlabeled hand dataset, called {\em RHTD}. The hand images in the RHTD are utilized to be real examples $y$. Some examples in RHTD are shown in Figure~\ref{fig:train_dataset_1}.
%

%% file: hpe_setting.tex
\section{Experimental Setting}

{\flushleft {\bf Hand Pose Estimators.}}
Two existing hand pose estimators are used to assess the quality of the synthesized hand pose images in our experiments: {\em Hand3D}~\cite{zimmermann_iccv17} and {\em convolutional pose machine} (CPM)~\cite{wei_cvpr16}. 
The two estimators infer the $3$D and $2$D hand poses from a monocular RGB image, respectively.
Both estimators are popular and often serve as the baselines for advanced estimators, such as \cite{cai_eccv18,mueller_cvpr18,panteleris_wacv18,simon_cvpr17,yuan_cvpr18}. 
Thus, if the synthetic hand pose images we generate can improve Hand3D and CPM, those images can also facilitate the follow-up research of Hand3D and CPM.

\begin{figure}[t]
    \begin{center}
    \begin{tabular}{ccc}
    		\bmvaHangBox{ 
    		\includegraphics[height=0.36in]{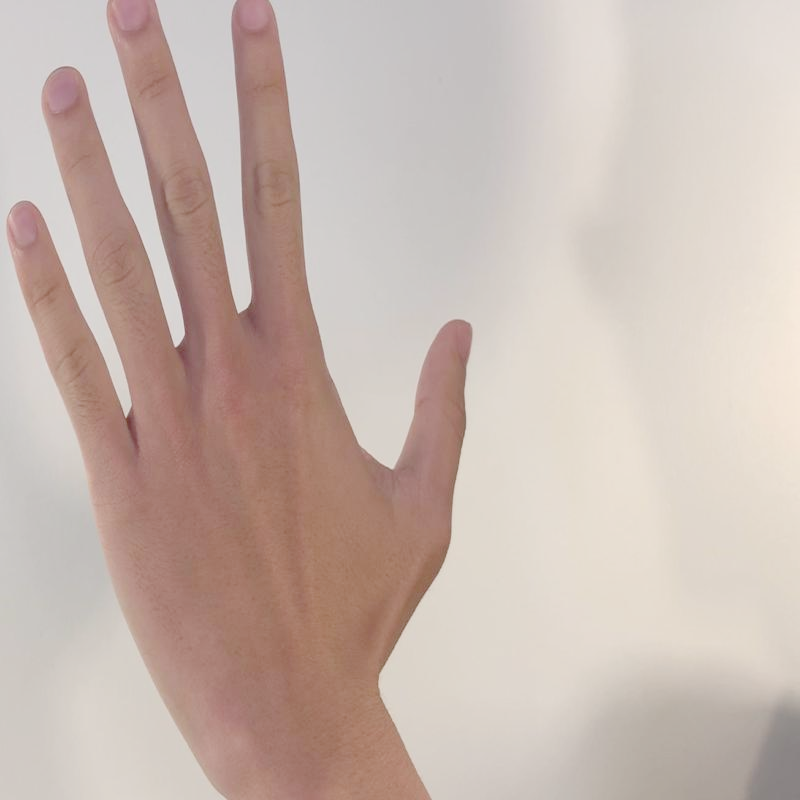}
    		\includegraphics[height=0.36in]{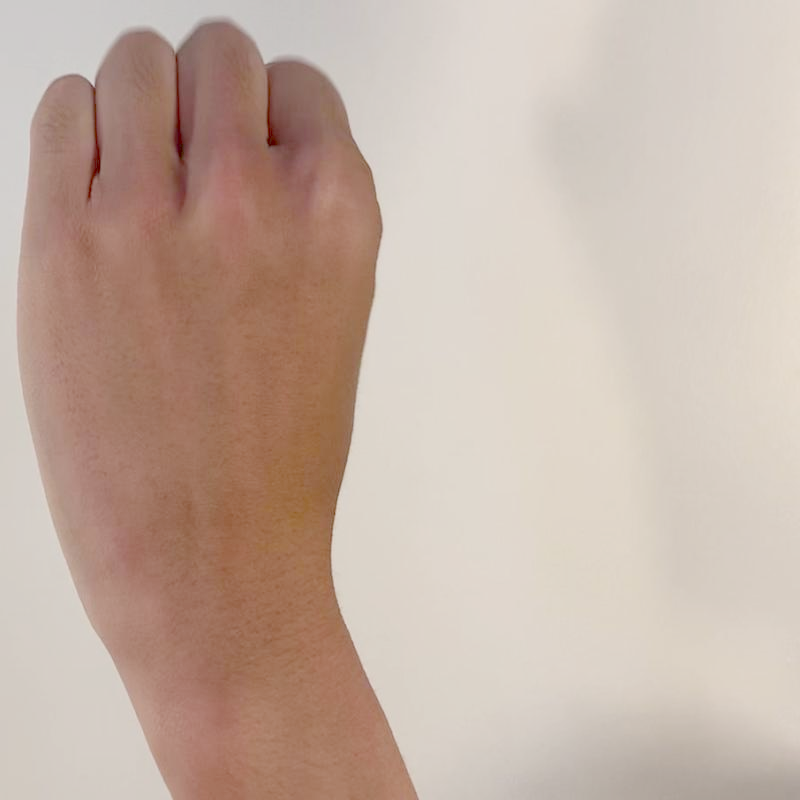}
    		\includegraphics[height=0.36in]{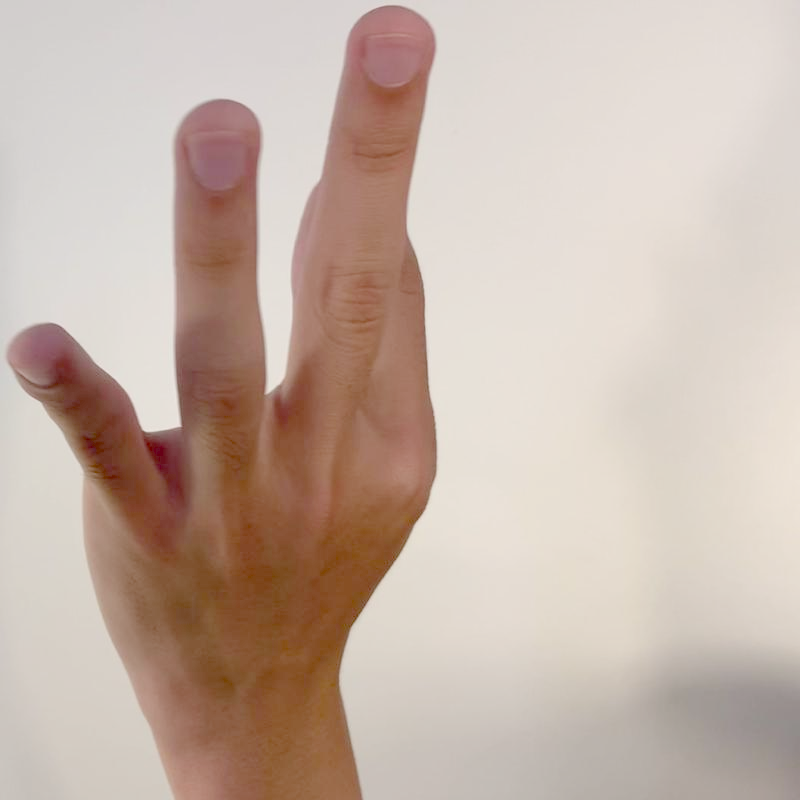}
    		\includegraphics[height=0.36in]{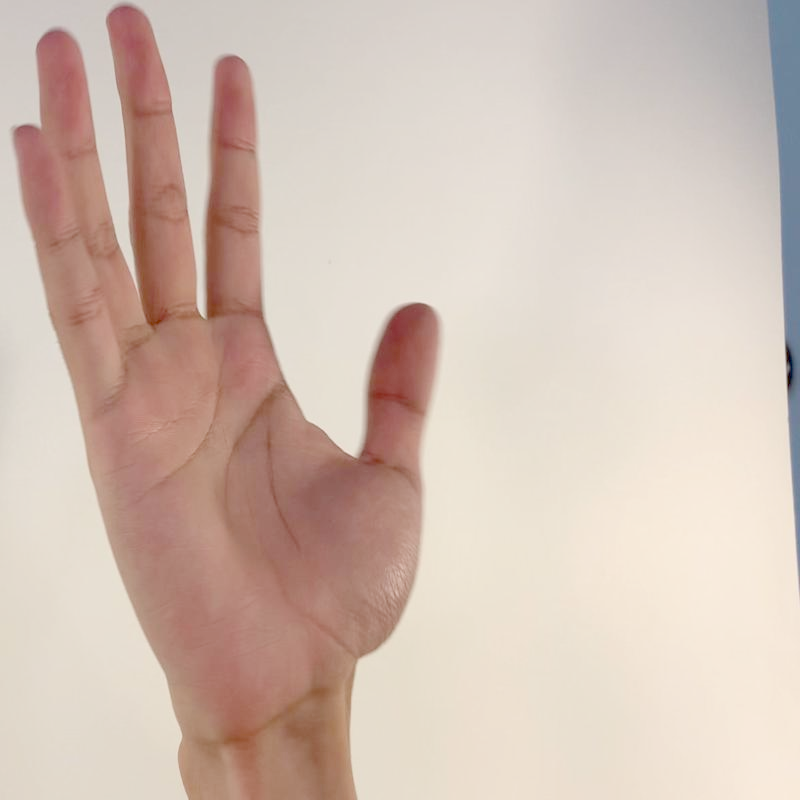}
    		\label{fig:train_dataset_1}}&
            \hspace{-0.2cm}
            \bmvaHangBox{
            \includegraphics[height=0.36in]{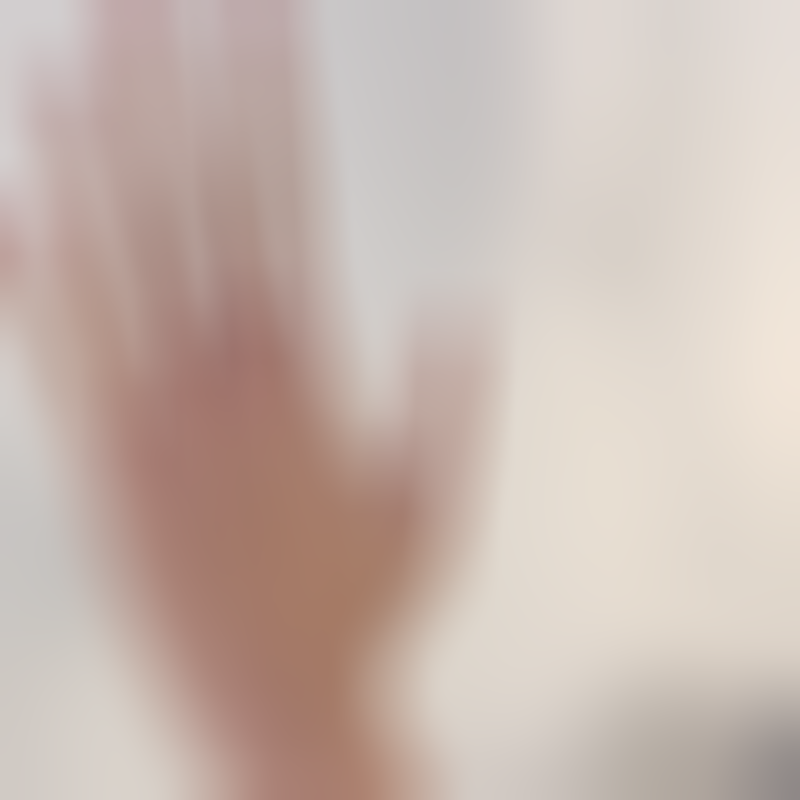}
            \includegraphics[height=0.36in]{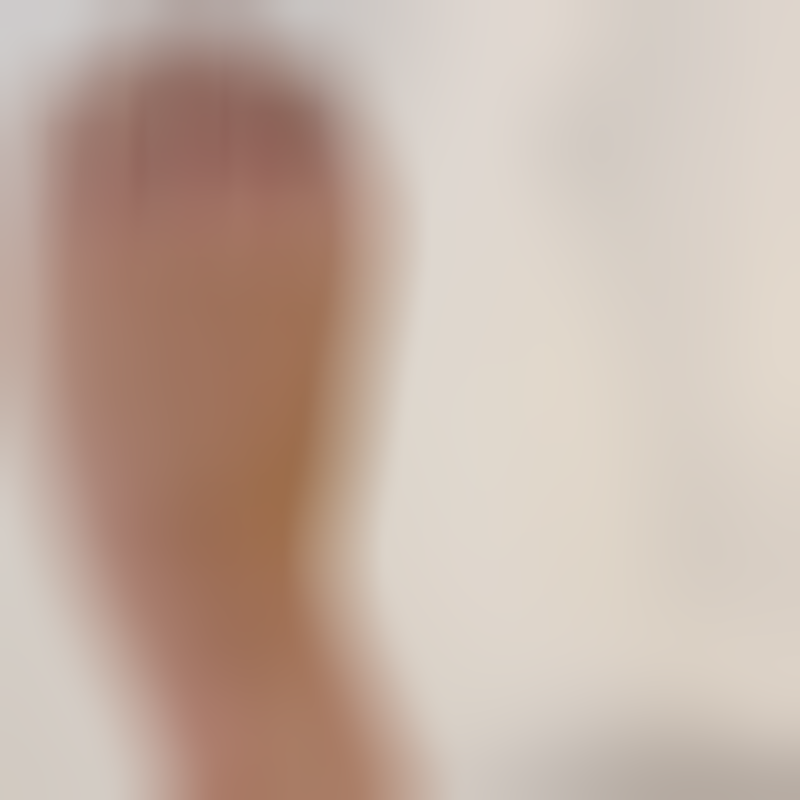}
    		\includegraphics[height=0.36in]{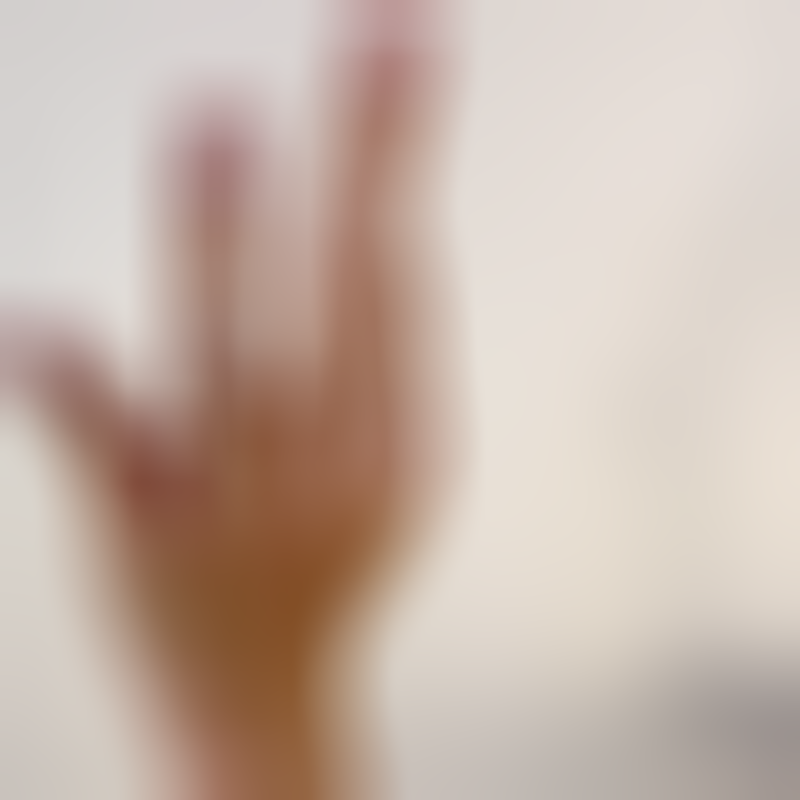}
    		\includegraphics[height=0.36in]{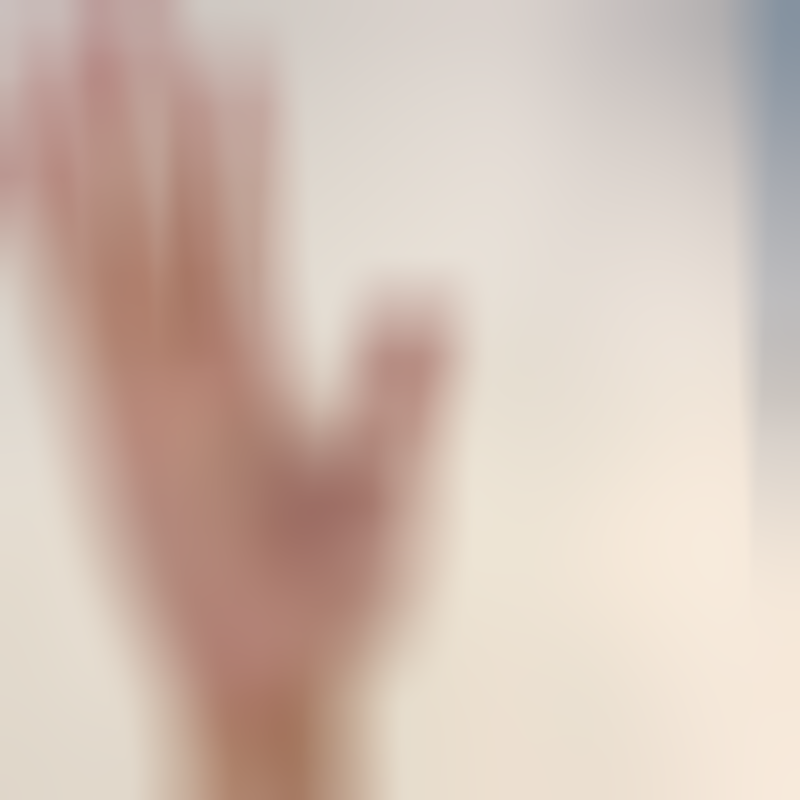}
    		\label{fig:train_dataset_2}}&
    		\hspace{-0.2cm}
            \bmvaHangBox{
           	\includegraphics[height=0.36in]{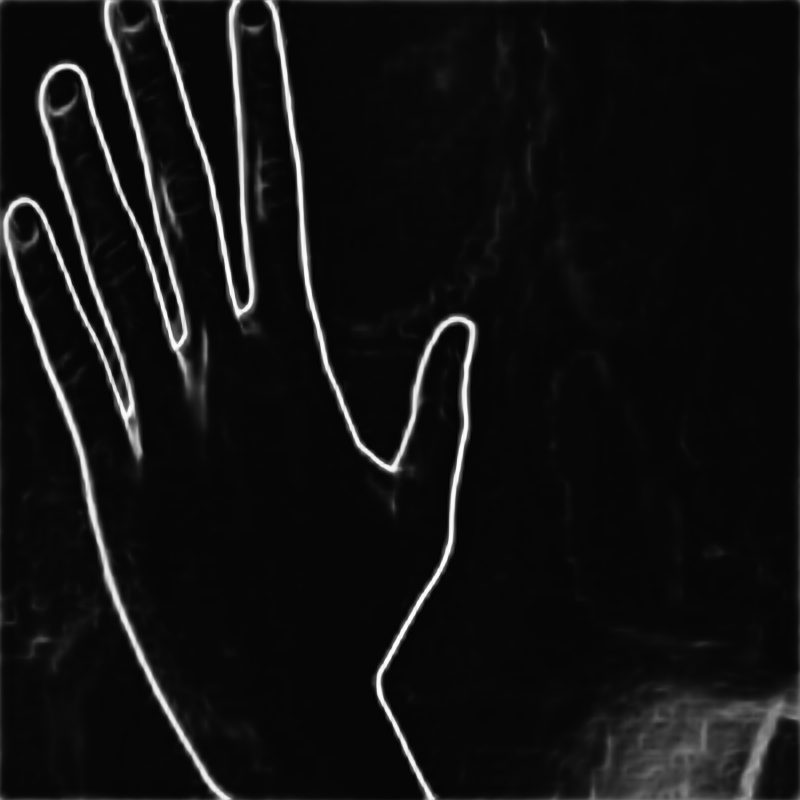}
    		\includegraphics[height=0.36in]{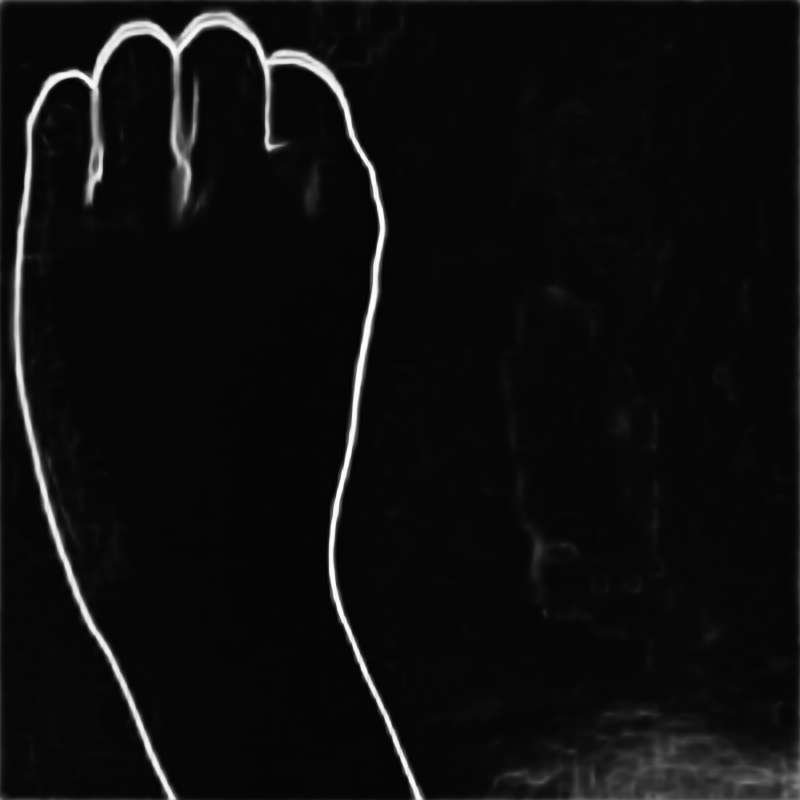}
    		\includegraphics[height=0.36in]{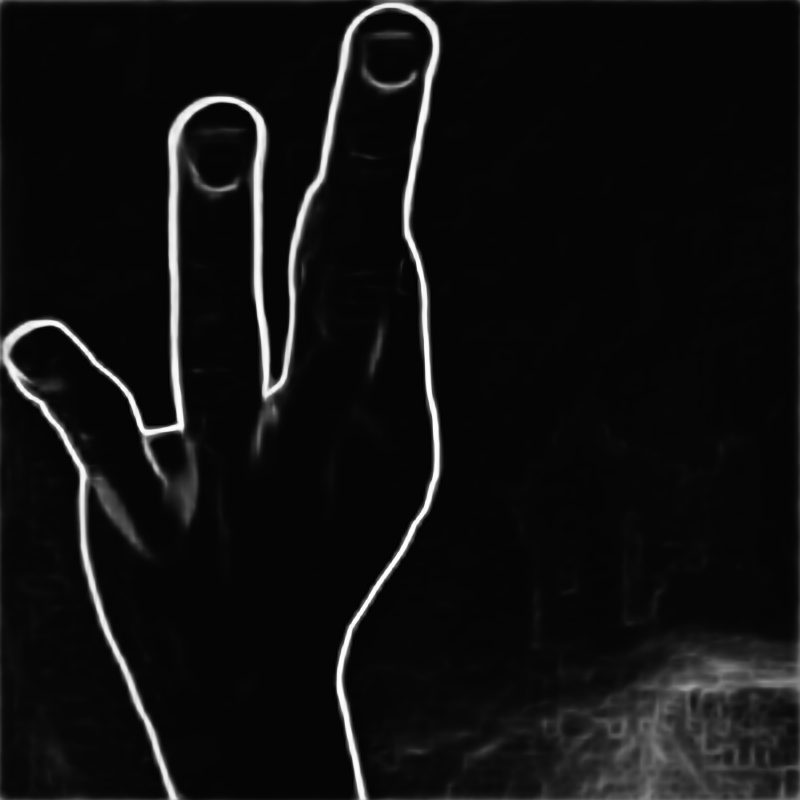}
    		\includegraphics[height=0.36in]{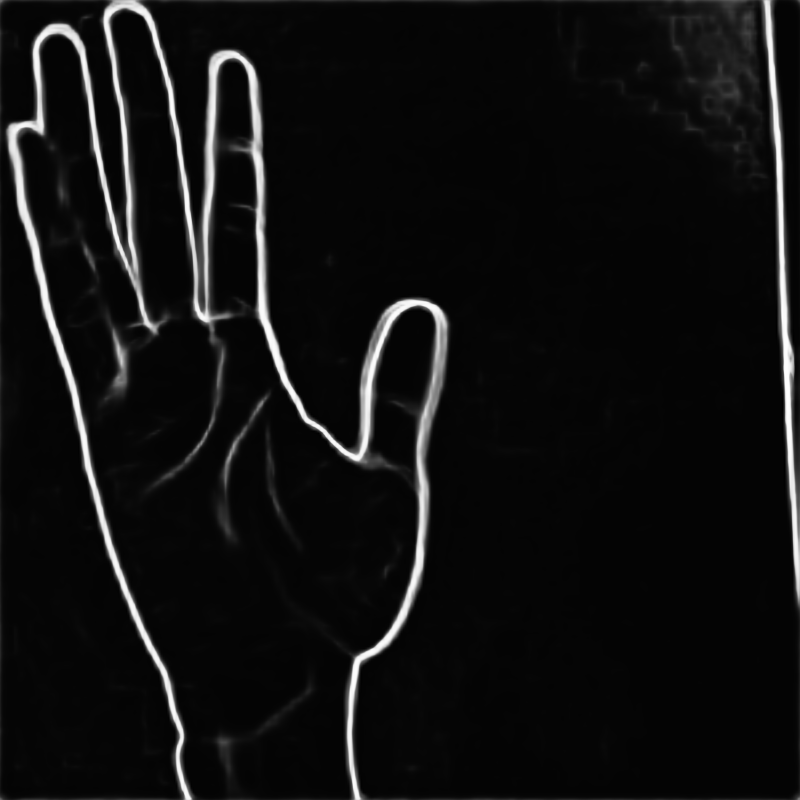}
    		\label{fig:train_dataset_3}}\\
    (a) & \hspace{-0.2cm} (b) & \hspace{-0.2cm} (c)
    \end{tabular}
	\end{center}
	\caption{Four examples of the RHTD dataset upon which TAGAN is learned. Each example is composed of (a) an image $x$, (b) its color map $x_b$ and (c) shape maps $x_s$.
	}
	\label{fig:train_dataset}
\end{figure}

\begin{figure}[t]
	\begin{center}
	\begin{tabular}{ccc}
    		\bmvaHangBox{
			\includegraphics[height=0.325in]{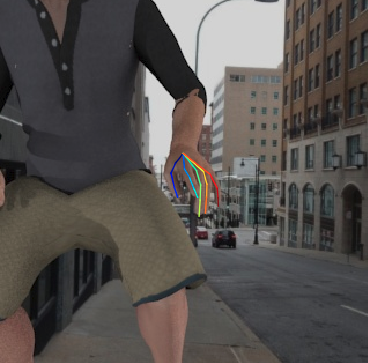}
    		\includegraphics[height=0.325in]{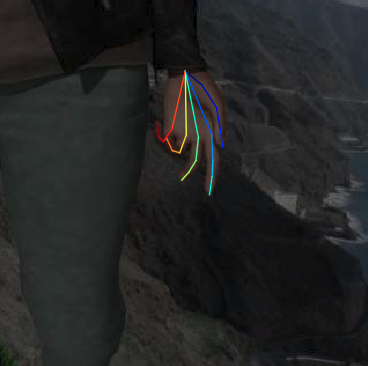}
    		\includegraphics[height=0.325in]{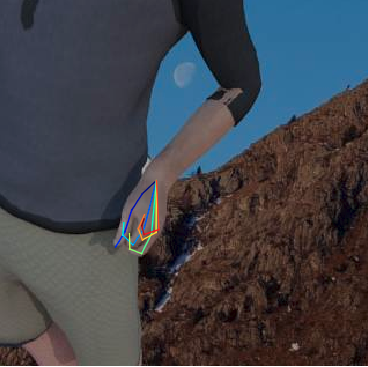}
    		\includegraphics[height=0.325in]{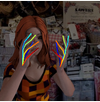}
    		\label{fig:rhd}}&
            \hspace{-0.2cm}
            \bmvaHangBox{
    		\includegraphics[height=0.325in]{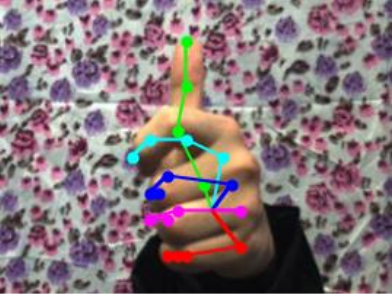}
    		\includegraphics[height=0.325in]{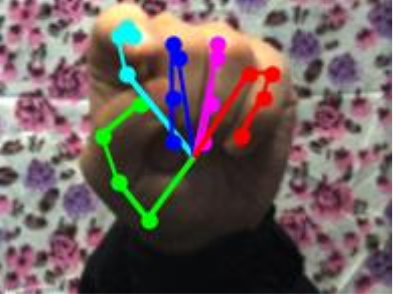}
    		\includegraphics[height=0.325in]{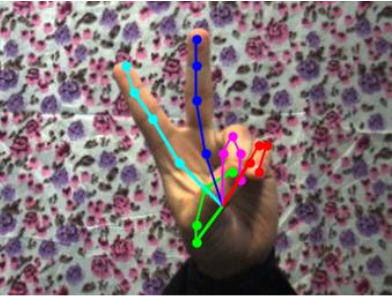}
    		\includegraphics[height=0.325in]{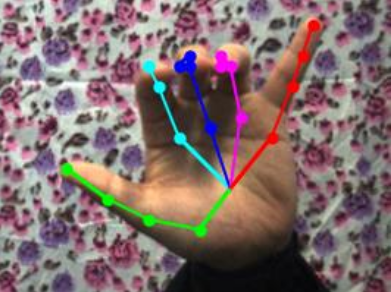}
    		\label{fig:stb}}&
    		\hspace{-0.2cm}
            \bmvaHangBox{
    		\includegraphics[height=0.325in]{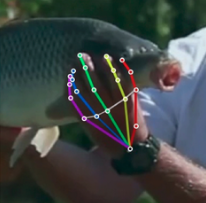}
    		\includegraphics[height=0.325in]{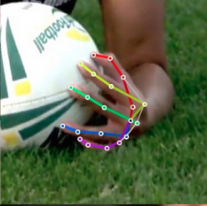}
    		\includegraphics[height=0.325in]{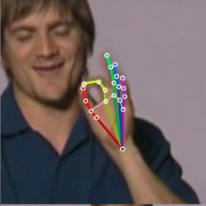}
    		\includegraphics[height=0.325in]{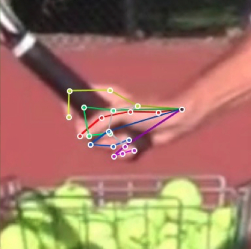}
    		\label{fig:cmu}}\\
    (a) & \hspace{-0.2cm} (b) & \hspace{-0.2cm} (c)
	\end{tabular}
	\end{center}
	\caption{Some examples in the three benchmark datasets. (a) The RHP dataset provides synthetic hand images with $3$D hand keypoints. (b)
	The STB dataset contains real hand images with $3$D keypoints. (c) The CMU-PS dataset offers real hand images with $2$D keypoints.}
	\label{fig:dataset}
\end{figure}


%
%

\begin{figure}[t] 
	\begin{center}
	\setlength\tabcolsep{0.5pt}
		\begin{tabular}{lcccccc} 
		

		
		
		\includegraphics[height=0.4in]{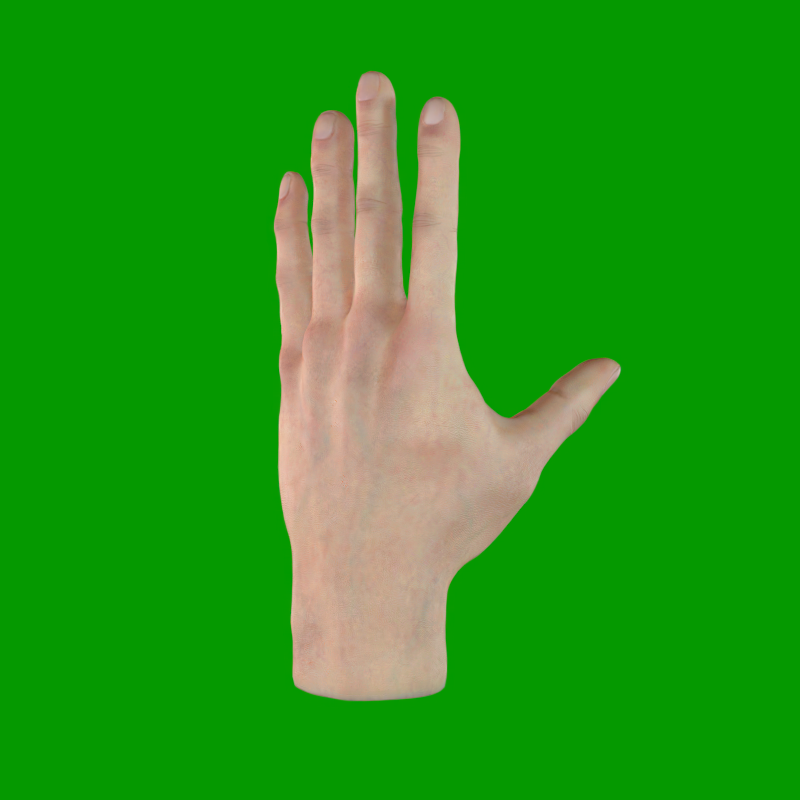}&
		\includegraphics[height=0.4in]{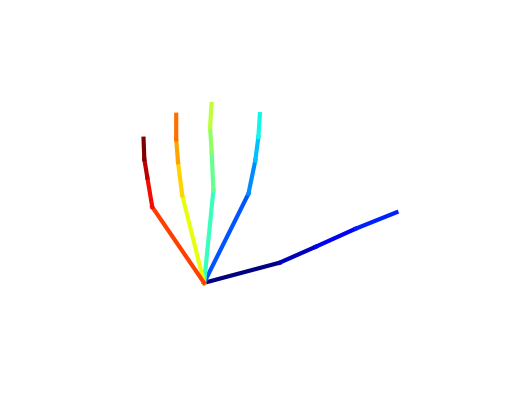} &
		\includegraphics[height=0.4in]{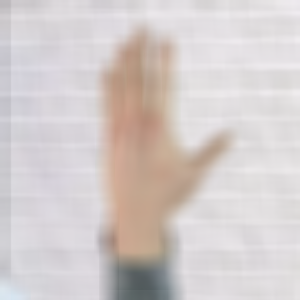} &
		\includegraphics[height=0.4in]{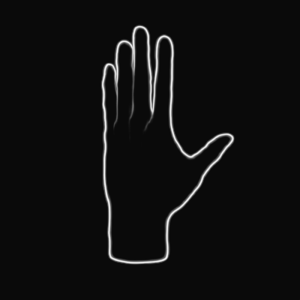} &
		\includegraphics[height=0.4in]{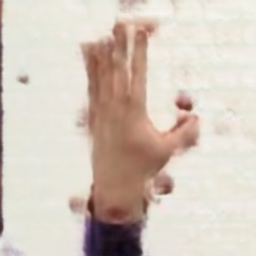} &
		\includegraphics[height=0.4in]{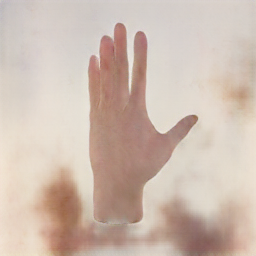} &
		\includegraphics[height=0.4in]{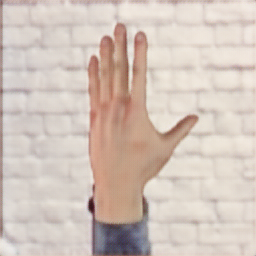} \\
    
        \includegraphics[height=0.4in]{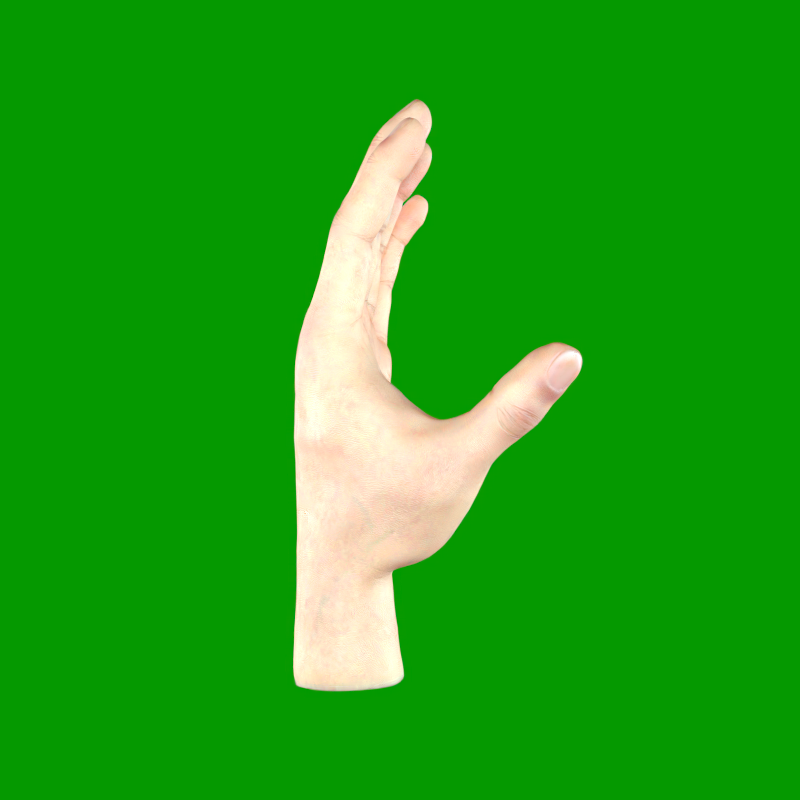}&
		\includegraphics[height=0.4in]{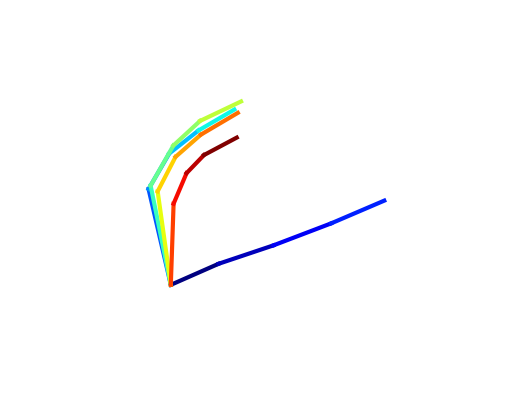} &
		\includegraphics[height=0.4in]{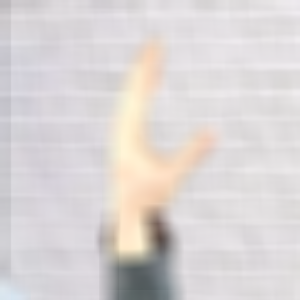} &
		\includegraphics[height=0.4in]{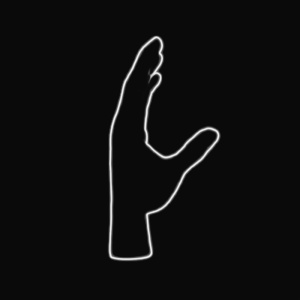} &
		\includegraphics[height=0.4in]{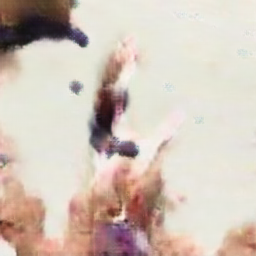} &
		\includegraphics[height=0.4in]{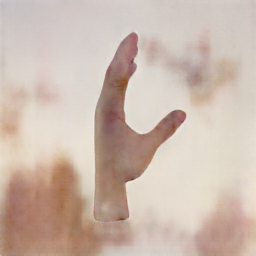} &
		\includegraphics[height=0.4in]{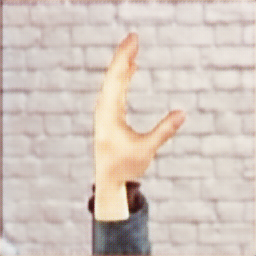} \\
        
         \includegraphics[height=0.4in]{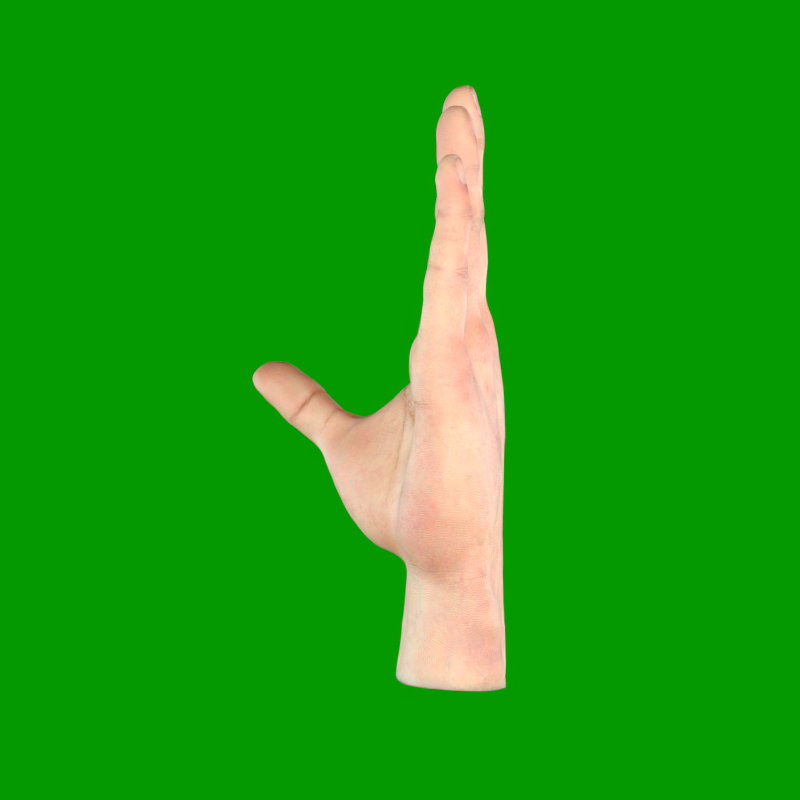}&
 		\includegraphics[height=0.4in]{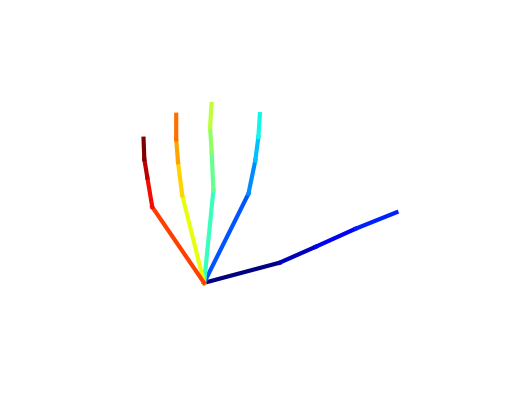} &
 		\includegraphics[height=0.4in]{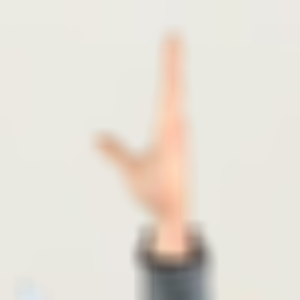} &
		\includegraphics[height=0.4in]{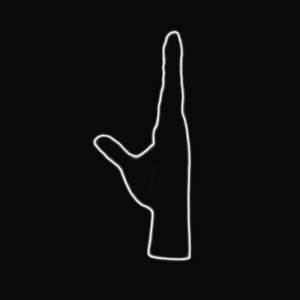} &
 		\includegraphics[height=0.4in]{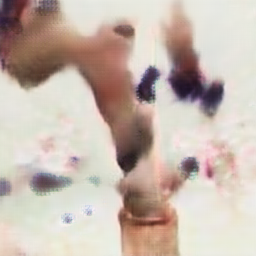} &
 		\includegraphics[height=0.4in]{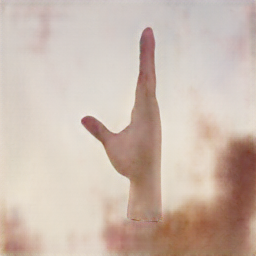} &
 		\includegraphics[height=0.4in]{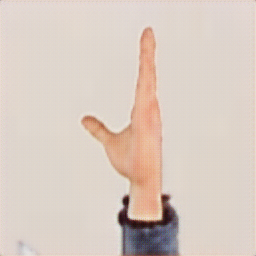} \\
        
        \includegraphics[height=0.43in]{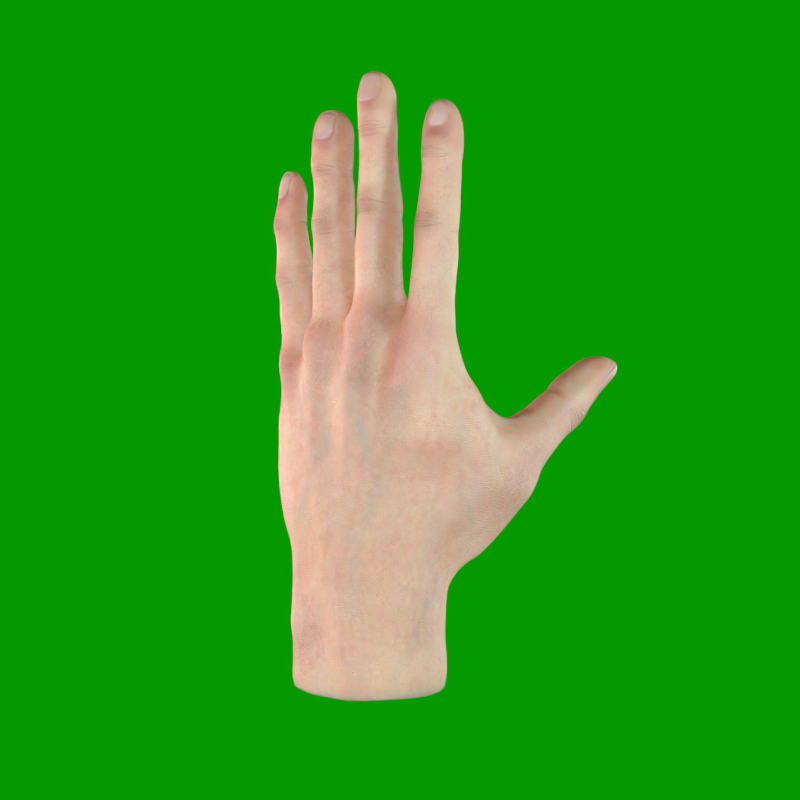}&
		\includegraphics[height=0.43in]{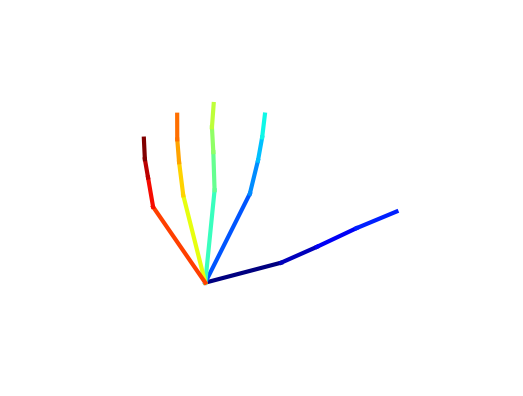} &
		\includegraphics[height=0.43in]{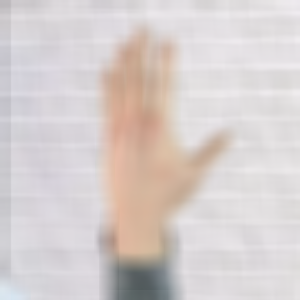} &
		\includegraphics[height=0.43in]{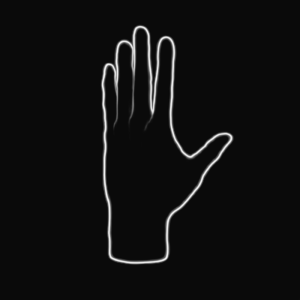} &
		\includegraphics[height=0.43in]{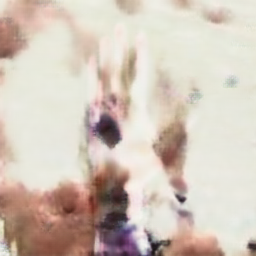} &
		\includegraphics[height=0.43in]{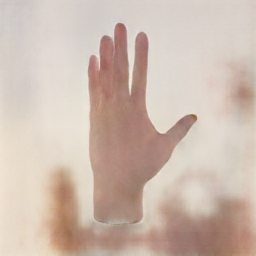} &
		\includegraphics[height=0.43in]{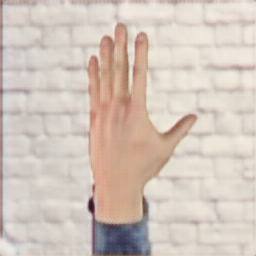} \\
		
		\includegraphics[height=0.43in]{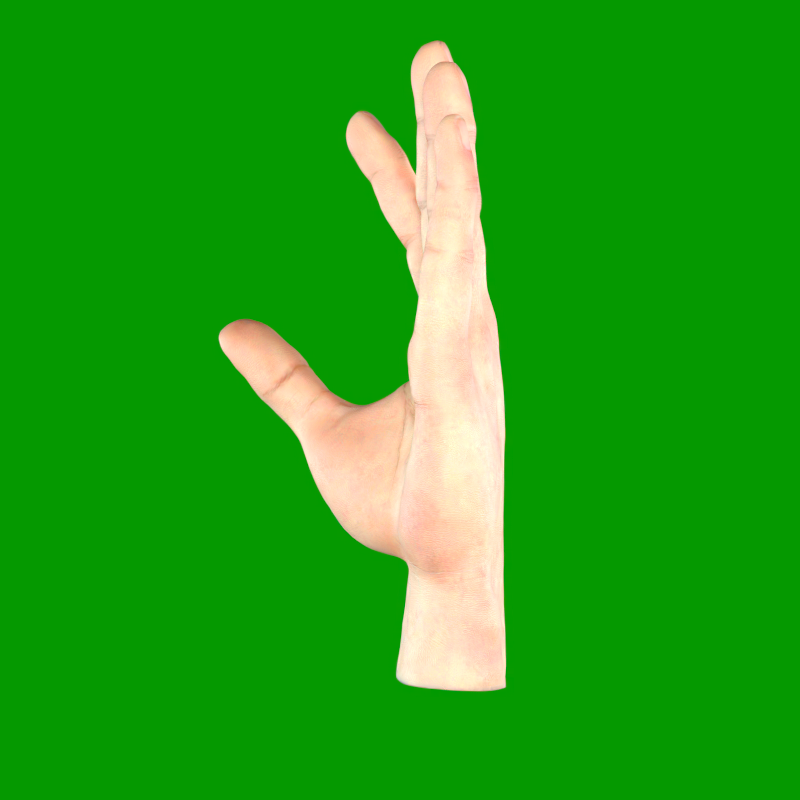}&
		\includegraphics[height=0.43in]{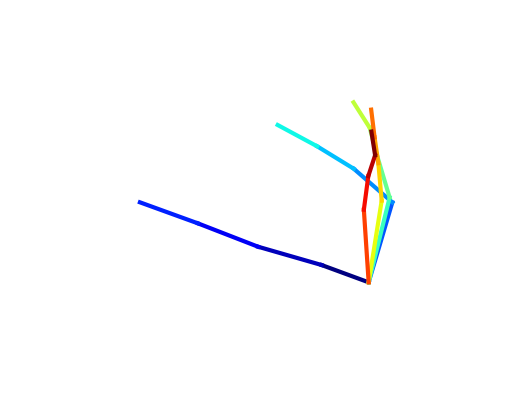} &
		\includegraphics[height=0.43in]{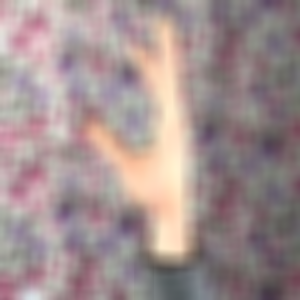} &
		\includegraphics[height=0.43in]{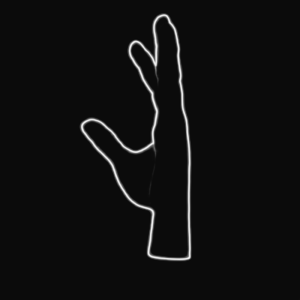} &
		\includegraphics[height=0.43in]{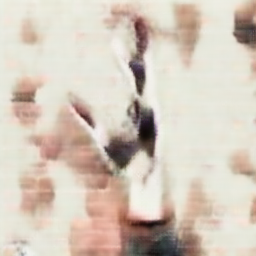} &
		\includegraphics[height=0.43in]{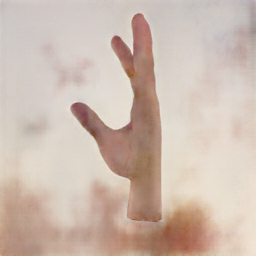} &
		\includegraphics[height=0.43in]{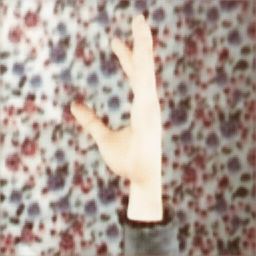} \\
		\end{tabular}
	\end{center}
	\caption{Comparison of the hand pose images synthesized by different methods. Images from left to right are 1) hand poses generated by the AR simulator, 2) their keypoint annotations, 3) color maps,  4) shape maps, the synthesized results by 5) CycleGAN~\cite{zhou_eccv18}, 6) Pix2pix Net~\cite{isola_cvpr17}, and 7) the proposed TAGAN, respectively. 
}
	\label{fig:comparison_generated_data}
\end{figure}

{\flushleft {\bf Dataset for Training.}}
To train the generators, including CycleGAN, pix2pix net, and the proposed TAGAN, for hand image synthesis, we collect $17,040$ real unlabeled hand images captured from people performing various hand gestures. 
This dataset is called the {\em real hand training dataset} (RHTD), which contains hand images with various poses, perspective views, and lighting conditions.
Some examples of RHTD are shown in Figure~\ref{fig:train_dataset}. 
To train the proposed TAGAN, images in RHTD come with the pre-computed edge~\cite{xie_cvpr15} and color maps.
In addition to RHTD, we adopt the AR simulator to generate $60,000$ synthetic hand images with various poses, perspectives, and lighting conditions. 
Some synthetic hand image examples are displayed in the first column of Figure~\ref{fig:comparison_generated_data}. 
By using our synthetic hand image generation process shown in Figure~\ref{fig:affine}, the synthetic hand images are then present at the appropriate locations of the real images (background).
The TAGAN is then applied to blend the synthetic hands with the real background images.

{\flushleft {\bf Datasets for Evaluation.}}
To evaluate the quality and the efficacy of the synthesized data, we select three benchmark datasets for evaluation including the {\em Rendered Hand Pose} (RHP)~\cite{zimmermann_iccv17}, {\em Stereo Tracking Benchmark} (STB)~\cite{zhang_arxiv16}, and {\em CMU Panoptic Studio} (CMU-PS)~\cite{simon_cvpr17} datasets.
Figure~\ref{fig:dataset} displays some examples of the three datasets.
%
The RHP dataset contains $41,258$ training and $2,728$ testing hand samples captured from $20$ subjects performing $39$ actions.
Each sample consists of an RGB image, a depth map, and the segmentation masks for the background, person, and each finger.
Each hand is annotated with its $21$ keypoints in both $2$D coordinates and $3$D world coordinate positions. 
The RHP dataset is split into a validation set (R-val) and a training set (R-train). 
%
%
%
The STB dataset provides $18,000$ hand images. It is split into two subsets: the stereo subset (STB-BB) and the color-depth subset (STB-SK). 
%
%
%
%
The CMU-PS dataset provides $1,912$ examples for training and $846$ examples for testing.
The $2$D hand keypoints of these examples are available.


{\flushleft {\bf Evaluation Metrics.}}
Following~\cite{cai_eccv18,simon_cvpr17,zimmermann_iccv17}, we adopt two metrics for evaluating the estimated hand poses, including the average {\em End-Point-Error} (EPE) and the {\em Area Under the Curve} (AUC) on the {\em Percentage of Correct Keypoints} (PCK).
We report the performance on both $2$D and $3$D hand pose estimation where the performance metrics are computed in pixels (px) and millimeters (mm), respectively. 
The performance of $3$D hand joint prediction is measured by using the PCK curves averaged over all $21$ keypoints. 
We use $2$D PCK and $3$D PCK to evaluate our approach on the RHP, STB, and CMU-PS datasets, respectively. 


%% file: hpe_results.tex
\section{Experimental Results}

This section evaluates the proposed TAGAN. 
First, we visually inspect the hand images generated by TAGAN and compare them with those generated via CycleGAN~\cite{zhou_eccv18} and Pix2pix Net~\cite{isola_cvpr17}. 
Second, we perform an ablation study to independently verify the efficacy of additional hand pose data and the importance of using TAGAN as a background-aware generator.  
Third, we show that the hand pose estimators can be greatly improved by fine-tuning with the data generated by TAGAN.


\subsection{Comparison of the Synthesized Hand Pose Images}
Figure~\ref{fig:comparison_generated_data} compares the quality of the generated hand images by using CycleGAN, Pix2pix Net, and the proposed TAGAN, respectively.
%
The CycleGAN learns one translation mapping from synthetic hand images to real images, and another mapping from real to synthetic images.
The Pix2pix Net derives the translation from a shape map to a realistic image.
TAGAN takes both foreground shapes and background tonality into account.
%
%
In Figure~\ref{fig:comparison_generated_data}, the first two columns show the synthetic hand images generated by the AR simulator and their hand-keypoint labels, respectively.
The third and the forth columns show the pre-computed color maps and shape maps from those synthetic hand images.
The last three columns display the results generated by CycleGAN, Pix2pix Net, and TAGAN, respectively.

As shown in Figure~\ref{fig:comparison_generated_data}, the CycleGAN does not have the shape and color constraints, so it tends to generate unnatural hand images.
Although the Pix2pix Net can successfully generate hand's shapes by using shape features, it does not consider the consistency of colors. 
The generated hand images still look unnatural.
The proposed TAGAN leverages real background information and gains color and shape constrains.
Hence, the generated results jointly maintain the color and shape features, making the synthesized images more realistic.

\begin{table}[t]
 \begin{minipage}[t]{.49\linewidth}
  \scriptsize
  \caption{$3$D pose estimation results.
  }
  \label{table:3d}
  \centering
    \begin{tabular}{p{2.9cm}ccc}
    \hline
 & {\scriptsize AUC}$\uparrow$& {\scriptsize EPE mean}$\downarrow$ \\ 
\hline
    {\scriptsize RHP}                            & 0.42         & 35.6\\
    {\scriptsize RHP$+$AR Hand w/ CBG}           &0.56        & 26.4\\
    {\scriptsize RHP+AR Hand  w/ RBG}         & 0.57        & 26.4\\
    {\scriptsize \bf{RHP+AR Hand w/ TAGAN}}     & \bf{0.60}    &\bf{24.2}\\
    \hline
    {\scriptsize STB}                             & 0.66          &15.7\\
    {\scriptsize STB$+$AR Hand w/ CBG}      & 0.67          &8.2\\
    {\scriptsize STB+AR Hand  w/ RBG}         & 0.71          &7.1\\
    {\scriptsize \bf{STB+AR Hand w/ TAGAN}}     & \bf{0.75}     &\bf{7.0}\\
    \hline
    \end{tabular}
 \end{minipage}
 \hfill
\hspace{-3mm}
 \begin{minipage}[t]{.49\linewidth}
  \scriptsize
    \caption{$2$D pose estimation results.
    }
    \hspace{-3mm}
  \label{table:2d}
  \centering
    \begin{tabular}{p{2.9cm}cc}
    \hline
    & {\scriptsize PCK@20} $\uparrow$           & {\scriptsize EPE mean} $\downarrow$\\ \hline
    {\scriptsize RHP}                             & 88.45             &10.76\\
    {\scriptsize RHP+AR Hand w/ RBG}         & 90.00             &9.80\\
    {\scriptsize \bf{RHP+AR Hand w/ TAGAN}}     & \bf{90.01}        &\bf{9.78}\\ 
    \hline
    {\scriptsize STB}                            & 96.6              &7.61\\
    {\scriptsize STB + AR Hand w/ RBG}         & 96.7              &7.59\\ 
    {\scriptsize \bf{STB + AR Hand w/ TAGAN}}     & \bf{97.0}         &\bf{7.55}\\  \hline 
    \end{tabular}
 \end{minipage}
\end{table}

\begin{figure}[t]
\begin{center}
\begin{tabular}{ccccc}
\includegraphics[height=0.4in]{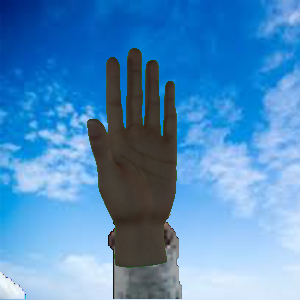}& 
\includegraphics[height=0.4in]{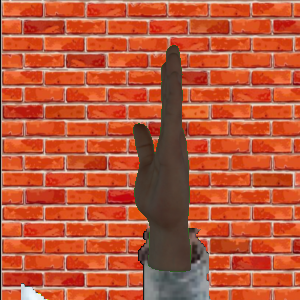}& 
\includegraphics[height=0.4in]{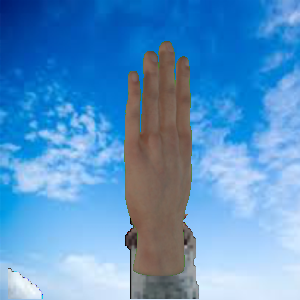}& 
\includegraphics[height=0.4in]{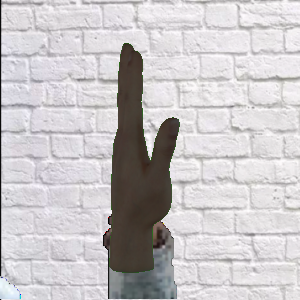}& 
\includegraphics[height=0.4in]{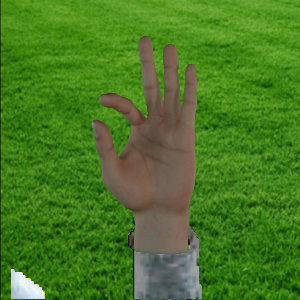} 
\end{tabular}
\end{center}
\vspace{-0.15in}
\caption{Tonality inconsistency in the ``AR Hand w/RBG'' data.}
\label{fig:woTAGAN}
\end{figure}

\subsection{Ablation Study}

For analyzing the pose estimator learning by using our generated data, we conduct an ablation study on both $2$D and $3$D hand pose estimation.
We adopt three comparative settings of training data on the STB datasets, including 
1) STB data with hand images produced by the AR simulator with clear background ``STB$+$AR hand w/ CBG",
2) STB data with AR hand images with real background ``STB$+$AR hand w/ RBG", 
and 3) STB with hand images generated by TAGAN ``STB$+$AR hand w/ TAGAN".
Some examples of the ``STB$+$AR hand w/ CBG'' and ``STB$+$AR hand w/ TAGAN" are shown in the first and the last columns of Figure~\ref{fig:comparison_generated_data}, respectively.
We train the hand pose estimators, Hand3D and CPM, on the training sets using the settings above, and test them on the validation sets, respectively. 
All the aforementioned experimental settings are also conducted on the RPH dataset. 

Table~\ref{table:3d} and Table~\ref{table:2d} show the experimental results on $3$D and $2$D hand pose estimation, respectively.
In EPE and PCK, we find that both $3$D and $2$D hand pose estimators can be greatly improved by using large-scale synthetic data.
%
%
%
%
Besides, Table~\ref{table:3d} and Table~\ref{table:2d} show that the pose estimators trained with AR hands and real background images can be further improved. 
The reason is that the augmented AR hand images with real backgrounds are closer to the real images.
Moreover, it can be observed that training hand pose estimators with data generated by TAGAN has higher PCK values and lower EPE errors than with the ``AR Hand w/ RBG'' data.

For $3$D pose estimation task, the pose estimator trained with ``AR Hand w/ CBG" is successfully improved in EPE mean $11.4$ ($=35.6-24.2$)mm and $8.7$ ($=15.7-7.0$)mm on the RHP and STB datasets, respectively, as shown in Table~\ref{table:3d}.
For $2$D pose estimation, the pose estimators are improved in EPE mean $0.98$($=10.76-9.78$) pixels and $0.06$($=7.61-7.55$) pixels on the RHP ans STB dataset, respectively, as shown in Table~\ref{table:2d}.
The quality of the data generated by TAGAN is better than the synthetic AR hand superimposed with real backgrounds (AR hands w/ RBG).
The reason is that the ``AR Hand w/ RBG'' data do not take the tonality consistency between synthetic hands and background into account. 
Some examples of tonality inconsistency are shown in Figure~\ref{fig:woTAGAN}.
To test the generation ability, we follow~\cite{simon_cvpr17} where hand pose estimators are trained on the STB training set and evaluated on the CMU-PS validation set. 
%
%
We adopt CPM in this experiment.
As shown in Table~\ref{table:2dcmu}, training the pose estimator with augmented hand images (AR hand w/ RBG) can enhance the performance. 
Moreover, training it with the data generated by TAGAN can gain more significant improvement in both PCK accuracy and EPE error.
The CPM is improved from $24.09$ to $28.81$ in PCK@20 accuracy and from $55.99$ to $52.39$ in EPE mean error.

\begin{table}[t]
\footnotesize
\caption{Results by training on STB and testing on CMU-PS.}
    \begin{center}
        \begin{tabular}{lcc}
        \hline
    &{\scriptsize PCK@20} $\uparrow$ &{\scriptsize EPE mean (mm)} $\downarrow$ \\ \hline
        {\scriptsize STB}                     & 24.09         & 55.99\\
        {\scriptsize STB+AR Hand w/ RBG}         & 24.82         & 56.67\\
        {\scriptsize \bf{STB+AR Hand w/ TAGAN}}     & \bf{28.81}    & \bf{52.93}\\ \hline
        \end{tabular}
        \end{center}
         \label{table:2dcmu}
\end{table}
\begin{figure}[t]
\begin{center}
\begin{tabular}{cc}
\bmvaHangBox{
\includegraphics[width=6.8cm]{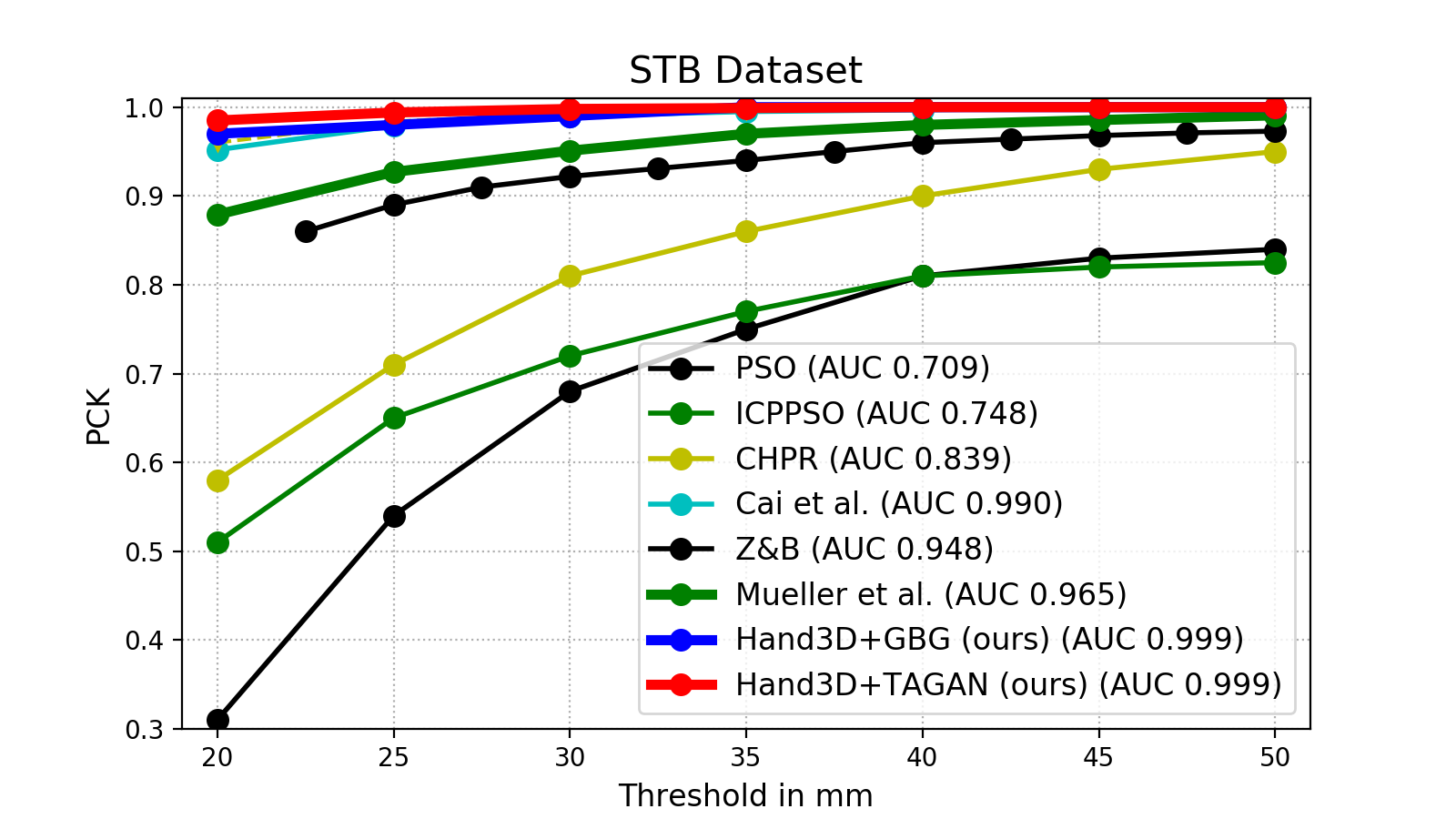}\label{fig:3DSTB}}&
\hspace{-0.35in}
\bmvaHangBox{
\includegraphics[width=6.8cm]{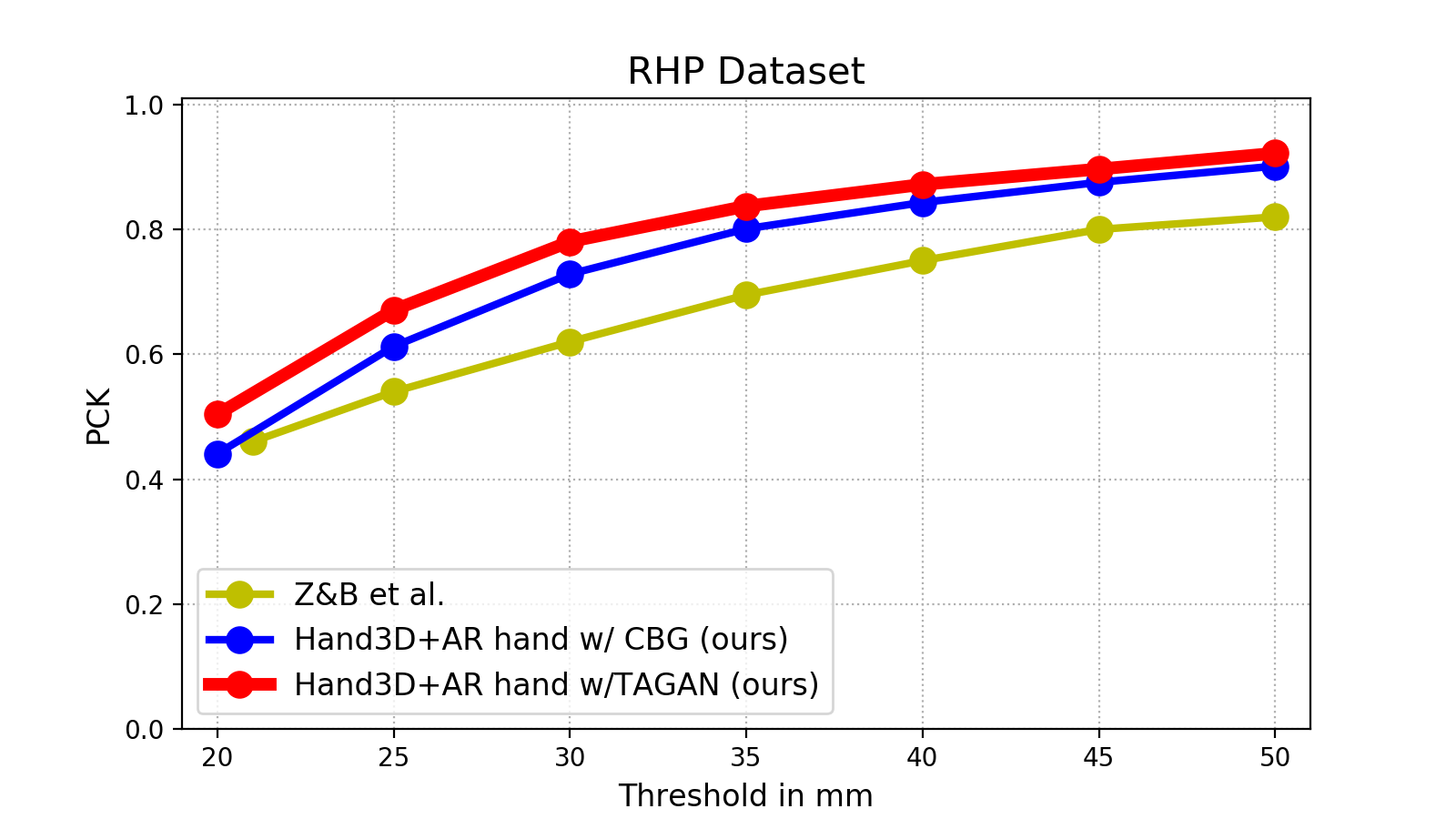}\label{fig:3DRHP}}\\
(a)&(b)\\ 
\end{tabular}
\end{center}
\caption{Comparison with the state-of-the-art approaches for $3$D pose estimation on the (a) STB and (b) RHP datasets.}
\label{fig:teaser}
\end{figure}

\subsection{Comparison with State-of-the-Art Results}

To compare with the existing $3$D hand pose estimators, we select the methods, PSO, ICPPSO, and CHPR as the baselines, and choose the state-of-the art methods including Cai~\etal in~\cite{cai_eccv18}, Z\&B in \cite{zimmermann_iccv17}, and Mueller~\etal in~\cite{mueller_cvpr18} for comparison on the STB dataset.
We select Z\&B \cite{zimmermann_iccv17} for comparison on the RHP dataset. 
We also provide the results by training our pose estimator with ``AR hand w/ RBG'' data for comparison.
Figure~\ref{fig:teaser} shows that training Hand3D with training data generated by TAGAN achieves the best performance.
To explore the improvement by training a $2$D/$3$D pose estimator with our generated data, we conduct two experiments on STB and RHP datasets.
We adopt Hand3D and CPM as the $2$D and $3$D pose estimators.
Table~\ref{table:3d} and~\ref{table:2d} show the results.  
We find that training either $2$D or $3$D pose estimators with the additional images generated by TAGAN reaches the best performance.


%% file: hpe_Conclusion.tex
\section{Conclusions and Future Work}
This study presents a novel data augmentation approach for improving hand pose estimation task.
To produce more realistic hand images for training pose estimators, we propose TAGAN, a conditional adversarial networks model, to blend the synthetic hand poses with real background images.
Our generated results align the hand shape and color tonality distribution between  synthetic hands and real background images.
The experimental results show that the state-of-the-arts hand pose estimators can be greatly improved with the aid of the training data generated by our method.
